\documentclass[10pt,twocolumn,letterpaper]{article}

\usepackage{wacv}              

\usepackage{graphicx}
\usepackage{amsmath}
\usepackage{amssymb}
\usepackage{booktabs}

\usepackage{hyperref}
\usepackage{url}
\usepackage{multirow}
\usepackage{graphicx}
\usepackage{enumitem}
\usepackage{amsmath,amsfonts,amssymb}

\usepackage{times}
\usepackage{epsfig}
\usepackage{graphicx}
\usepackage{amsmath}
\usepackage{amssymb}
\usepackage{multirow}
\usepackage{graphicx}
\usepackage{algorithm}
\usepackage{algpseudocode}
\usepackage{booktabs}
\usepackage{wrapfig}
\usepackage{caption}
\usepackage{url}
\usepackage[T1]{fontenc}

\newcommand{\tabref}[1]{Tab~\ref{#1}}

\usepackage[capitalize]{cleveref}
\crefname{section}{Sec.}{Secs.}
\Crefname{section}{Section}{Sections}
\Crefname{table}{Table}{Tables}
\crefname{table}{Tab.}{Tabs.}

\def\wacvPaperID{2274} 
\def\confName{WACV}
\def\confYear{2025}

\begin{document}

\title{Masked Autoregressive Model for Weather Forecasting}

\author{Doyi Kim\thanks{D.-Y. Kim and M.-S. Seo provided equal contributions to this work.}, Minseok Seo\textsuperscript{*}, Hakjin Lee, Junghoon Seo\\
SI Analytics\\
{\tt\small \{doyikim, minseok.seo, hakjinlee, jhseo\}@si-analytics.ai}
}

\maketitle

\begin{abstract}
The growing impact of global climate change amplifies the need for accurate and reliable weather forecasting. Traditional autoregressive approaches, while effective for temporal modeling, suffer from error accumulation in long-term prediction tasks. The lead time embedding method has been suggested to address this issue, but it struggles to maintain crucial correlations in atmospheric events.
To overcome these challenges, we propose the Masked Autoregressive Model for Weather Forecasting (MAM4WF). This model leverages masked modeling, where portions of the input data are masked during training, allowing the model to learn robust spatiotemporal relationships by reconstructing the missing information. MAM4WF combines the advantages of both autoregressive and lead time embedding methods, offering flexibility in lead time modeling while iteratively integrating predictions.
We evaluate MAM4WF across weather, climate forecasting, and video frame prediction datasets, demonstrating superior performance on five test datasets.
\end{abstract}

\section{Introduction}
\label{sec:intro}
The atmosphere on Earth is a chaotic system~\cite{lorenz1963deterministic}, making it challenging to accurately predict future states of weather phenomena, such as clouds, temperature, and other key variables.
Numerical Weather Prediction (NWP) was developed to solve this, relying on the laws of physics to generate forecasts.
However, despite decades of development and widespread use, NWP still faces challenges, including high computational costs and limited understanding of atmospheric chaos~\cite{bauer2015quiet}.
Recently, data-driven approaches to weather forecasting have shown great potential, rivaling operational NWP models~\cite{lam2022graphcast, pathak2022fourcastnet, keisler2022forecasting, gao2022earthformer, bi2023accurate}.
Many of these methods focus on autoregressive architectures due to their flexibility in forecasting various lead times~\cite{lam2022graphcast, pathak2022fourcastnet, keisler2022forecasting}.
However, these approaches often suffer from error propagation, as they use their own predictions as inputs for the next forecasts~\cite{bi2023accurate,ning2023mimo}.
To expand lead time options, several studies have exploited the lead time embedding method~\cite{sonderby2020metnet, espeholt2022deep}, which generates predictions based on a given lead time.
However, this approach may fail to maintain the spatiotemporal correlations necessary to understand weather phenomena accurately.

To address these limitations, we propose a new forecasting approach that combines the lead time embedding method with autoregressive modeling.
\begin{figure}[t!]
    \centering
    \includegraphics[width=1.0\columnwidth]{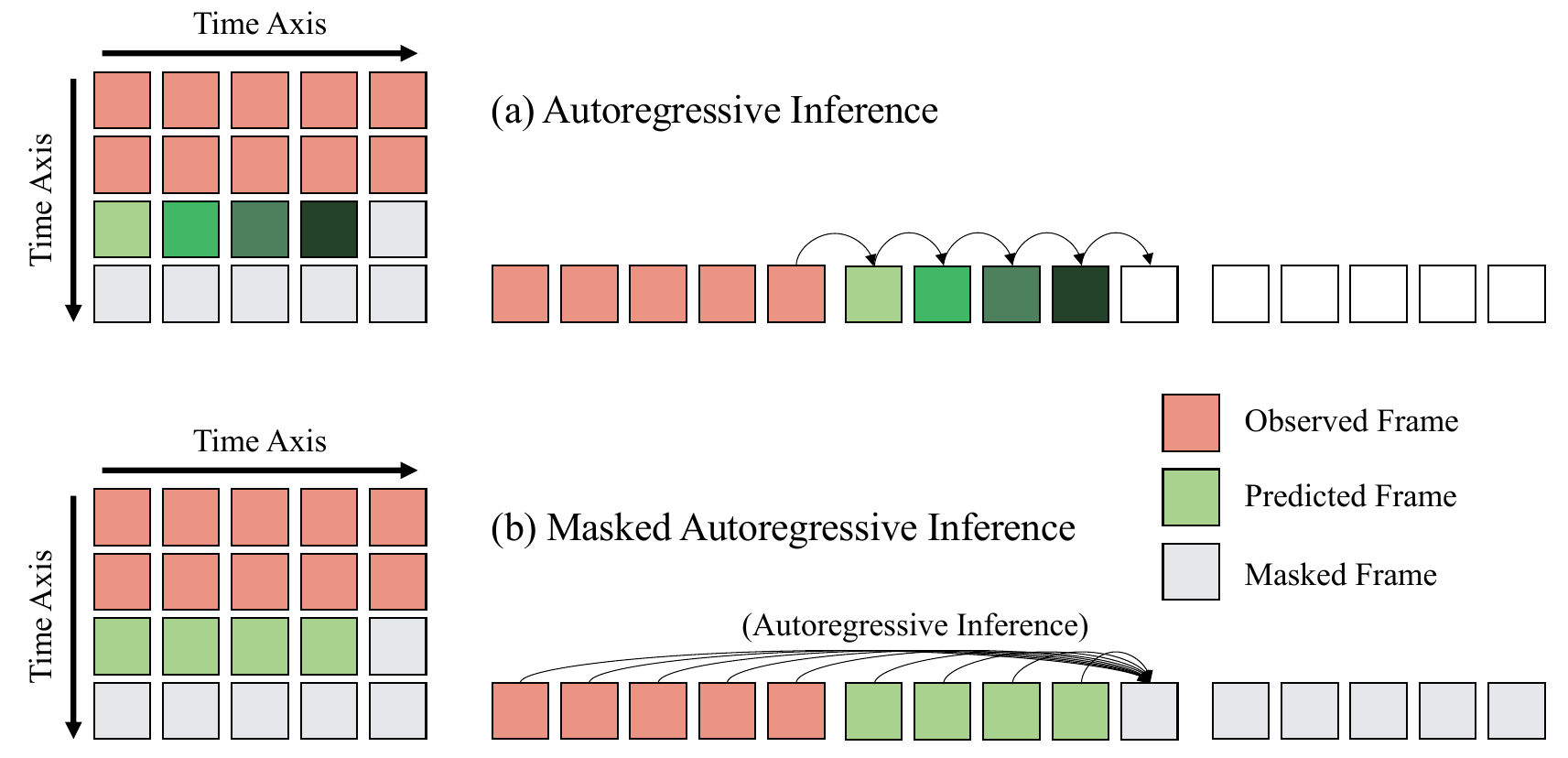}
    \caption{Comparison of (a) autoregressive and (b) masked autoregressive inference methods. (a) repeatedly uses frames with the trained input length for the next prediction. (b) is a structure that flexibly adjusts the input length using masked frames for the next prediction step. 
    }
    \label{fig:teaser}
\end{figure}
\begin{table*}[t!]
{
\begin{center}
\renewcommand{\arraystretch}{1.2}
\resizebox{1.8\columnwidth}{!}{%
\begin{tabular}{l|cc|cccccccccc|c}
\hline \hline
\multicolumn{1}{c|}{\multirow{2}{*}{\textbf{Method}}} & \multirow{2}{*}{\textbf{\#Param. (M)}} & \multirow{2}{*}{\textbf{Training Epoch}} & \multicolumn{10}{c|}{\textbf{Time Step}} & \multirow{2}{*}{\textbf{MSE}} \\ \cline{4-13}
\multicolumn{1}{c|}{}                                 &                                        &                                          & 1  & 2  & 3 & 4 & 5 & 6 & 7 & 8 & 9 & 10 &                               \\ \hline
SimVP-S (MIMO)                                        &20.4                                       & 2K                                         & 11.6   & 14.7   & 17.5  & 19.6  & 22.0  & 24.4  &  27.1 & 29.4  & 32.6  &  36.5  &  23.5                             \\
SimVP-S (MIMO)*10                                     & 20.4                                       & 20K                                         & 8.8    & 11.9   & 15.6  & 17.3  & 19.2  & 21.4  & 25.7  & 28.8   & 30.3  & 35.2   & 21.4                              \\
SimVP-L (MIMO)                                        &   53.5                                     &  2K                                        &  13.7  & 17.5   & 18.1  & 20.2 &  24.1 &  26.6 &  29.8 &  33.1 &  36.5 & 38.2   &   25.7                            \\ \hline
SimVP-S (MISO-Multi Model)                            & 20.4*10                                       &      20K                                    & \textbf{8.3}   &  \textbf{10.9}  & \textbf{13.1}  & \textbf{15.6}  &   \textbf{17.8}&  \textbf{20.0} & \textbf{22.4}  &  \textbf{24.5} & \textbf{26.1}  & \textbf{28.7}   &     \textbf{18.7}                          \\
SimVP-S (MISO-Autoregressive)                         &  20.4                                      &     2K                                     &  8.3   & 13.4   &  19.2 &  24.5 & 30.3  &36.2   & 42.6   & 48.7   &  55.2 &  62.3  & 34.1                                \\ \hline \hline
\end{tabular}%
}
\end{center}

\caption{This results from comparing the efficiency and accuracy of autoregressive and non-autoregressive models. The original SimVP model are modified with Multiple-In-Single-Out Multi Model (MISO-Multi) and Multiple-In-Single-Out Autoregressive (MISO-Autoregressive) structures. Except for the MISO-Autoregressive model, other models are tested in a non-autoregressive manner. All experiments were conducted on the Moving MNIST dataset.} 

\label{tab:motivation}
}
\end{table*}
Our Masked Autoregressive Model for Weather Forecasting (MAM4WF) retains the flexibility of the autoregressive model while avoiding error accumulation.  
Simultaneously, MAM4WF resolves weak correlations that appear in the output of the lead time embedding method using masked autoregressive techniques.
As shown in ~\cref{fig:teaser}-(b), MAM4WF uses unmasked inputs (observed frame) to generate predictions for future states.
We then propose a masked autoregressive method that reuses the observed and predicted frames to predict the next state. 
This iterative process helps make accurate and flexible predictions in tasks with multiple lead times.
MAM4WF incorporates both powerful data-driven techniques and achieves state-of-the-art results on two benchmark weather and climate prediction datasets.

Our contributions can be summarized as follows:

\begin{itemize}
\setlength\itemsep{0em}
\item We propose a novel masked autoregressive model that predicts future frames recursively with a Multiple-In-Single-Out (MISO) design that can consider the spatiotemporal correlation of weather events.
\item We propose a masked structure to capture both spatiotemporal correlations between input data and output data.
\item We demonstrate that our masked autoregressive Model for Weather Forecasting (MAM4WF) achieves state-of-the-art performance on two benchmark datasets for weather and climate prediction.
\end{itemize}

\section{Related Work}
\label{sec:relwork}
Weather forecasting and video prediction share a common challenge: predicting future states from past sequences. Two primary approaches have emerged in this domain: future frame prediction~\cite{shi2015convolutional, lu2017flexible, yu2019crevnet, wang2019eidetic, wang2019memory, castrejon2019improved, su2020convolutional, guen2020disentangling, wu2021motionrnn, chang2021mau, wang2022predrnn, gao2022simvp, chang2022strpm} and future frame generation~\cite{xu2018video, denton2018stochastic, babaeizadeh2018stochastic, franceschi2020stochastic, akan2021slamp, voleti2022mcvd}. In weather forecasting, the focus is typically on future frame prediction, which estimates the most likely sequence based on past data~\cite{xu2018video, denton2018stochastic, babaeizadeh2018stochastic, franceschi2020stochastic, akan2021slamp, voleti2022mcvd}.
\begin{figure*}[t!]
    \centering
    \includegraphics[width=1.5\columnwidth]{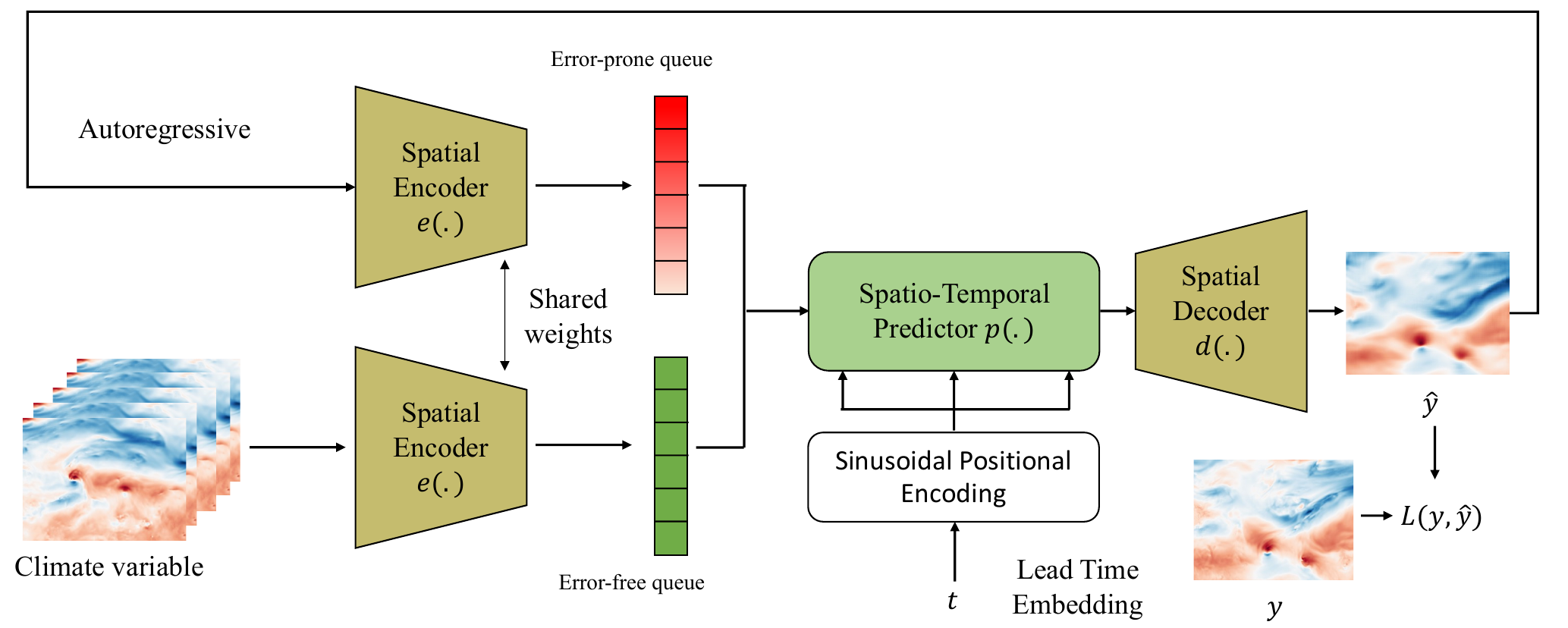}
    \caption{An overview of MAM4WF training process. MAM4WF is an implicit model for time step $t$. MAM4WF consists of a spatial encoder $e(\cdot)$, a spatio-temporal predictor $p(\cdot)$, and a spatial decoder $d(\cdot)$. Note that, MAM4WF is trained masked autoregressive manner with error-prone queue.}
    \label{fig:divmodnet}
\end{figure*}
\paragraph{Autoregressive Models} Autoregressive models have been widely applied in weather forecasting due to their ability to capture temporal dependencies. These models predict future states by using previous outputs as inputs for the next prediction. Early works, such as ConvLSTM\cite{shi2015convolutional} and RainNet\cite{ayzel2020rainnet}, have successfully combined autoregressive techniques with architectures like LSTM and U-Net. These models incorporate memory units, such as Long Short-Term Memory (LSTM)\cite{graves2012long} and Gated Recurrent Units (GRU)\cite{cho-etal-2014-learning}, which help store historical information and improve multi-step forecasting.

However, a major drawback of autoregressive models is error accumulation. Since each future prediction is based on the previous one, any errors propagate through the forecast, leading to significant degradation in accuracy over long-term predictions~\cite{wang2022predrnn}. This issue is particularly well-known for extended lead-time forecasts, where forecast uncertainty increases over time.
\paragraph{Lead Time Embedding} To mitigate the problem of error accumulation inherent in autoregressive methods, the lead time embedding method has been proposed.
This method incorporates the target lead time directly into the model as an input, allowing it to generate predictions for specific time horizons without relying on previously generated outputs~\cite{sonderby2020metnet}. This approach reduces the accumulation of errors over time by avoiding the need for recursive predictions.

A prominent example of this approach is MetNet\cite{sonderby2020metnet}, which processes spatiotemporal data along with the desired lead time to create tailored predictions.
By focusing on lead time as an input feature, the model can bypass the challenges associated with autoregressive techniques. Espeholt \textit{et al. (2022)} extended this method by integrating physics-based constraints into the model, which further improves its ability to make accurate predictions\cite{espeholt2022deep}.

Despite these advancements, the lead time embedding method has limitations. By structure, it does not utilize feedback from previous forecasts, which can reduce long-term correlation among predicted outputs.
This lack of sequential feedback can impair the ability to maintain consistency in long-range forecasts.
\paragraph{Non-Autoregressive Models} There is also an alternative approach that predicts all future frames simultaneously rather than sequentially.
This strategy solves the error accumulation problem, making it particularly attractive for long-term forecasting tasks. Models like Deep Voxel Flow\cite{liu2017video} and FutureGAN\cite{aigner2018futuregan} use spatiotemporal convolutional networks to model the relationships between frames without relying on previous outputs. More recent models, such as SimVP\cite{gao2022simvp} and MIMO-VP\cite{ning2023mimo}, have demonstrated that non-autoregressive methods can achieve state-of-the-art results by predicting multiple frames in parallel.

The key advantage is avoiding cumulative errors, as predictions are generated all at once.
For example, SimVP uses a Multiple-In-Multiple-Out (MIMO) strategy to capture temporal relationships across observed frames, achieving strong performance with a relatively simple CNN-based architecture. Similarly, MIMO-VP extends transformer architectures with local spatiotemporal blocks, allowing efficient and accurate prediction across multiple benchmarks~\cite{ning2023mimo}.

However, non-autoregressive models are not without their weaknesses. This method also predicts multiple outputs, which makes it difficult to capture correlations between the predicted values. It is a significant problem in weather forecasting, where time continuity is considered important. 
%
\section{Motivation}
The limitation of autoregressive models in weather forecasting is error accumulation.
%
On the other hand, non-autoregressive models offer a promising alternative by predicting multiple frames at once (MIMO approach), avoiding the sequential error propagation of autoregressive models.
%
However, we question whether the MIMO architecture is essential for achieving high performance, or if the improvements are primarily driven by the use of multiple inputs.
To explore this, we compared autoregressive and non-autoregressive models with the same structure and hyperparameters based on a modified SimVP MIMO model.

Along with MIMO, we experimented with two variations of the MISO model, the MISO-Autoregressive and MISO-Multi Model, to evaluate the impact of multiple inputs and outputs.
Our results, shown in Table~\ref{tab:motivation}, indicate that while multi-input models benefit from additional inputs, the MISO-Multi Model performs best.
This demonstrates that multiple inputs, not multiple outputs, are key to improving predictions.
Despite this, the MISO-Autoregressive model suffers from long-term error accumulation, which underscores the need for a model that combines the strengths of multiple input structures and autoregressive approaches while addressing error propagation.

Thus, we propose a new weather forecasting model based on the following principles:

\begin{itemize}
    \item \textbf{Multiple-In-Single-Out}: Using multiple inputs to capture spatiotemporal dynamics for accurate forecasting.
    \item \textbf{Single End-to-End Model}: Ensuring computational efficiency by using a single model instead of multiple ones.
\end{itemize}

This approach aims to develop a robust and efficient model that minimizes the challenges of error accumulation while ensuring accurate long-term weather predictions.
\section{Masked Autoregressive Model for Weather Forecasting}
\paragraph{Model Design} In our model framework named Masked Autoregressive Model for Weather Forecasting (MAM4WF), the goal of the future frame prediction model \( F(\cdot) \) is to map the input \( X \in \mathbb{R}^{C \times T \times H \times W} \) to the target output \( y_t \in \mathbb{R}^{C \times H \times W} \) using the lead time \(t\) and the history of model predictions \(\hat{Y}_{t-1}\).
The learnable parameters of the model are denoted by \( \Theta \). These parameters are optimized to minimize the following objective function:
\begin{equation}
\min_{\Theta} \sum_{t < \hat{T}} L(F_{\Theta}(X, \hat{Y}_{t-1},t), y_{t}),
\end{equation}
where the objective is to learn the mapping from inputs to the \(\hat{T}\)-time future frame \(y_{t}\).
In this context, \(X = \{x_{T-1}, x_{T-2} , \ldots, x_{0}\}\) represents the \(T\) observed frames, \( \hat{Y}_{t}=\{\hat{y}_{1}, \hat{y}_{2}, \ldots, \hat{y}_{t}\}\) represents the history of \(t\) predicted frames, \(Y = \{y_{1}, y_{2}, \ldots, y_{\hat{T}}\}\) represents the \(\hat{T}\) ground-truth of future frames, \(C\) is the number of channels, \(T\) is the observed frame length, \(\hat{T}\) is the future frame length, \(H\) is the height, and \(W\) is the width of the frames. We define \(\hat{Y}_{0}\) as the empty set, with this case being an exception.
Various loss functions \(L\) can be utilized for optimization, such as mean squared error (MSE)~\cite{gao2022simvp}, mean absolute error (MAE)~\cite{ning2023mimo}, smooth loss~\cite{seo2022domain}, or perceptual loss~\cite{shouno2020photo}.
In this work, we specifically employ the MSE loss, which is the most commonly used loss function for future frame prediction tasks.
\subsection{Model Instantiation}
~\cref{fig:divmodnet} illustrates our model instantiation of MAM4WF in this work.
MAM4WF comprises multiple modules, including an encoder, error-free and error-prone queues, a predictor, and a decoder.
Given an input \( X \) and target time step \( t \), \( X \) passes through the encoder and its output is given to the error-free queue. 
The iterative prediction process up to the \( t-1 \) time step, $ \hat{Y}_{t-1}=\{\hat{y}_{1}, \hat{y}_{2}, \ldots, \hat{y}_{t-1}\} $, is stacked in the error-prone queue.
Subsequently, the error-free queue and error-prone queue are concatenated, and the concatenated output is passed through the predictor and decoder to produce the final output $\hat{y}_{t}$.

\paragraph{Encoder-Predictor-Decoder Framework}
In the task of video frame prediction, the current leading architectural paradigm is the \textit{encoder-predictor-decoder} structure~\cite{gao2022simvp}.
Unlike autoregressive models, the encoder-predictor-decoder structure is trained using convolutional neural networks (CNN) to map multi-step inputs $X$ to multi-step outputs $Y$. 
The \textit{encoder} $e(\cdot)$ functions to extract features from observed frames $X$.
In contrast to the \textit{encoder}, which treats the observed frames as independent images without considering their spatiotemporal relationships, the \textit{predictor} $p(\cdot)$ is tasked with capturing these relationships and transforming them into features for predicting future frames.
Finally, the \textit{decoder} $d(\cdot)$ reconstructs the forthcoming frames $Y$ based on the features provided by the \textit{predictor}.
MAM4WF adopts the encoder-predictor-decoder structure, which consists of (Conv, LayerNorm~\cite{ba2016layer}, SiLU~\cite{elfwing2018sigmoid}) for the encoder, (Conv, LayerNorm, SiLU, PixelShuffle~\cite{shi2016real}) for the decoder, and ConvNeXt~\cite{liu2022convnet} blocks for the predictor.
In contrast to the original encoder-predictor-decoder structure, our model output is a specific future frame corresponding to the given target lead time, rather than predicting multiple future frames simultaneously.

\begin{table*}[t!]
\begin{center}
\resizebox{1.7\columnwidth}{!}{%
\begin{tabular}{l|ccc|ccc|ccc}
\hline
\hline
            & \multicolumn{3}{c|}{\textbf{Moving MNIST}} & \multicolumn{3}{c|}{\textbf{TrafficBJ}} & \multicolumn{3}{c}{\textbf{Human 3.6}} \\ \hline
\textbf{Method}      & MSE       & MAE       & SSIM      & MSE $\times 100$      & MAE      & SSIM     & MSE$ / 10$      & MAE $ / 100$    & SSIM     \\ \cline{1-10} 
ConvLSTM~\cite{shi2015convolutional} &   103.3        &  182.9         &   0.707        &  48.5        &   17.7       &   0.978       &  50.4        &  18.9       &    0.776      \\
PredRNN~\cite{wang2017predrnn}       &   56.8     &     126.1      &      0.867        &    46.4      &    17.1      &    0.971      &   48.4       &    18.9     &     0.781     \\
Causal LSTM~\cite{wang2018predrnn++} &   46.5      &    106.8       &    0.898        &     44.8     &     16.9     &     0.977     &    45.8      &     17.2    &      0.851    \\
MIM~\cite{wang2019memory}            &   44.2    &     101.1      &       0.910        &        42.9  &    16.6      &    0.971      &   42.9       &    17.8     &    0.790      \\
E3D-LSTM~\cite{wang2018eidetic}      &   41.3    &      86.4     &       0.920        &     43.2     &      16.9    &     0.979     &    46.4      &     16.6    &     0.869     \\
PhyDNet~\cite{guen2020disentangling} &   24.4      &   70.3        &   0.947      &        41.9  &       16.2   &         0.982 &        36.9  &       16.2  &        0.901  \\
SimVP~\cite{gao2022simvp}            &   23.8      &      68.9     &     0.948             & 41.4         & 16.2         &0.982          & 31.6         & 15.1        &   0.904       \\
MIMO-VP*~\cite{ning2023mimo}        &   17.7      &       51.6    &     0.964             & -        & -         & -          & -         & -        &   -       \\ \hline
MAM4WF      &    \textbf{15.3}       &     \textbf{49.2}      &    \textbf{0.966}       &    \textbf{37.2}     &     \textbf{16.4}     &      \textbf{0.983}           & \textbf{12.6}     &       \textbf{11.2}   &   \textbf{0.942}       \\ \hline \hline
\end{tabular}
}

\end{center}
\caption{Performance comparison results of MAM4WF and recent leading approaches on three future frame prediction common benchmark datasets. MAM4WF achieved state-of-the-art performance on all three benchmark datasets, which have different characteristics: Moving MNIST, TrafficBJ, and Human 3.6. The asterisk (*) indicates the performance reported in the author's paper. See Supplementary Material Section 4 for qualitative comparison results.}
\label{tab:common}
\end{table*}
\paragraph{Error-prone Queue \& Error-free Queue}
To ensure that MAM4WF retains the history of the initial observation and predictions, it incorporates two key components: an \textit{error-free queue} and an \textit{error-prone queue}.
The error-free queue \(Q_{\text{error-free}}\) serves as an explicit memory bank, storing feature vectors derived from the observed frames \(X\). This explicit memory queue enables the model to preserve all the information from the initial observation when making forecasts across all lead times.
In contrast, common autoregressive models undergo alterations of initial observed information as the lead time increases.
In addition to this, MAM4WF introduces an error-prone queue \(Q_{\text{error-prone}}\), which comprises feature vectors from the history of predicted frames \(\hat{Y}\).
This component explicitly maintains the history of predictions and enables the model to consider spatiotemporal correlations between predictions, which is not guaranteed in the common lead time embedding approaches. 
\paragraph{Lead Time Embedding}

The approach in MAM4WF utilizes the lead time embedding methodology for flexible lead time prediction. For a given lead time \( t \), sinusoidal positional embedding is performed at position \( t \). This embedded representation, \( t_{\text{embed}} \), is subsequently passed through an MLP, where its dimensionality is adjusted to match the number of channels in each layer of the predictor \( p(\cdot) \).

The sinusoidal positional encoding for a given position \( t \) and dimension \( i \) is articulated as proposed by \cite{vaswani2017attention}:
\begin{align}
    PE_{(t, 2i)} &= \sin\left(\frac{t}{{10000^{\frac{2i}{d}}}}\right), \\
    PE_{(t, 2i+1)} &= \cos\left(\frac{t}{{10000^{\frac{2i}{d}}}}\right),
\end{align}
where \( PE \) denotes the 2D positional encoding matrix, \( i \) is the dimension index, and \( d \) stands for the embedding dimension. The complete positional embedding for \( t \) is obtained by aggregating across the dimension \( i \):
\begin{equation}
    t_{\text{embed}} = \left[PE_{(t, 0)}, PE_{(t, 1)}, \ldots, PE_{(t, d-1)}\right].
\end{equation}
Upon computation of the embedding \( t_{\text{embed}} \), it is forwarded through two fully-connected layers equipped with the GELU activation function \cite{hendrycks2016gaussian}: {(Linear, GELU, Linear)}. In the context of this work, the lead time embedding \( s(\cdot) \) is described as a composition of the sinusoidal positional embedding and processing through a two-layer network.

\subsection{Learned Prior (\textit{LP})}
Although randomly masking the future queue improves the model robustness to incomplete future queues, the feature vectors extracted from predicted frames still do not accurately reflect those from actual video frames.
As a result, prediction errors may increase as the time step progresses during the testing phase. To address this, we introduce a \textit{Latent Projection (LP)} that aims to project feature vectors from predicted results closer to those from actual video frames.
As shown in ~\cref{fig:teaser}, \textit{LP} takes the model output $\hat{y}{t}$ from $F(\cdot)$ and generates corresponding feature vectors. It is trained to minimize the MSE loss between the feature generated by $e(y{t})$ and the feature generated by \textit{LP}.
Consequently, \textit{LP} produces features that are more similar to the ground truth, even in the presence of errors in the predicted results.
~\cref{fig:teaser} illustrates MAM4VP at inference time. Note that \textit{LP} is trained through fine-tuning.
\begin{table*}[t!]
\begin{center}
\resizebox{1.9\columnwidth}{!}{%
\begin{tabular}{c|cccccc|ccccc}
\hline \hline
\multirow{2}{*}{\textbf{Model}} & \multicolumn{6}{c|}{\textbf{SEVIR}}                                 & \multicolumn{5}{c}{\textbf{ICAR-ENSO}}                              \\ \cline{2-12} 
                       & \#Param. (M) & GFLOPS & CSI-M  & CSI-160 & CSI-16 & MSE    & \#Param. (M) & GFLOPS & C-Niño3.4-M & C-Niño3.4-WM & MSE   \\ \hline 
UNet                   & 16.6         & 33     & 0.3593 & 0.1278  & 0.6047 & 4.1119 & 12.1         & 0.4    & 0.6926      & 2.102        & 2.868 \\
ConvLSTM               & 14.0         & 527   & 0.4185 & 0.2157  & 0.7441 & 3.7532 & 14.0         & 11.1   & 0.6955      & 2.107        & 2.657 \\
PredRNN                & 23.8         & 328   & 0.4080 & 0.2928  & 0.7569 & 3.9014 & 23.8         & 85.8   & 0.6492      & 1.910        & 3.044 \\
PhyDNet                & 3.1          & 701    & 0.3940 & 0.2767  & 0.7507 & 4.8165 & 3.1          & 5.7    & 0.6646      & 1.965        & 2.708 \\
E3D-LSTM               & 12.9         & 523   & 0.4038 & 0.2708  & 0.7059 & 4.1702 & 12.9         & 99.8   & 0.7040      & 2.125        & 3.095 \\
Rainformer             & 19.2         & 170    & 0.3661 & 0.2675  & 0.7573 & 4.0272 & 19.2         & 1.3    & 0.7106      & 2.153        & 3.043 \\
Earthformer            & 7.6          & 257   & 0.4419 & 0.3232  & 0.7513 & 3.6957 & 7.6          & 23.9   & 0.7329      & 2.259        & 2.546 \\ \hline
MAM4WF          &           34.7   &   392     &   \textbf{0.4607}     &     \textbf{0.3430}    &      \textbf{0.7761}  &       \textbf{2.9371} &     34.2         &   11.8     &   \textbf{ 0.7698}         &       \textbf{2.484}       &  \textbf{1.563}     \\ \hline \hline
\end{tabular}%
}
\end{center}
\caption{Performance comparison experiment results between MAM4VP and recent leading weather prediction models on SEVIR and ICRA-ENSO datasets. Note that many-to-many models, such as MAM4VP, could not be tested on the weather prediction benchmark dataset because the model structure must be significantly changed when the input $T$ and output $\hat{T}$ are different.}
\label{tab:weather}
\end{table*}
\subsection{Model Training}
The masked autoregressive model \( F(\cdot) \) is trained to predict the future frame \( y_{t} \) by taking as input the observed frames \( X \), the sequentially predicted frames \( \hat{Y}_{t-1} \), and the lead time step \( t \).
Thus, given a pair of input frames \(X\), target frames \(Y\), and the final lead time \(\hat{T}\), objective function of the model \(F(\cdot)\) is computed as:
\begin{equation}
   \sum_{t < \hat{T}} L \left( d \left( p ( e(X), e(\hat{Y}_{t-1}), s(t) ) \right), y_{t} \right) ,
  \label{eq:object}
\end{equation}
where $e(\cdot)$ is the encoder, $p(\cdot)$ is the predictor, $d(\cdot)$ is the decoder, and $s(\cdot)$ is the lead time embedding.
%

%



The proposed model is trained sequentially, with \( Q_{\text{error-prone}} \) being accumulated from \( t=0 \) to \( \hat{T}-1 \).
It is noteworthy that \( Q_{\text{error-prone}} \) is pre-allocated as a zero tensor, following a similar approach as that used in Masked Autoencoders (MAE)~\cite{he2022masked}.
This training strategy enables both the predictor \( p(\cdot) \) and the decoder \( d(\cdot) \) to be trained on inputs that incorporate errors from \( Q_{\text{error-prone}} \) with the error-free features \( Q_{\text{error-free}} \) during training.
As a result, the predictions are robust against error accumulation during inference.
Furthermore, this approach facilitates the consideration of correlations between outputs.

\section{Experiments}
In this section, we present the evaluation of MAM4WF on widely used benchmarks for future frame prediction and weather$\slash$climate forecasting, respectively.
Additionally, we conducted an ablation study to gain insights into the design of video frame prediction models.
Due to the page limitation, some implementation details and additional experiment results are included in the Supplementary Material.
\paragraph{Training \& Test Datasets}
We evaluate MAM4VP using five datasets, as summarized in ~\tabref{tab:dataset}. Moving MNIST~\cite{srivastava2015unsupervised}, TrafficBJ~\cite{zhang2017deep}, and Human 3.6~\cite{ionescu2013human3} are commonly used benchmark datasets for future frame prediction. Moving MNIST consists of synthetically generated video sequences featuring two digits moving between 0 and 9, while TrafficBJ is a collection of taxicab GPS data and meteorological data recorded in Beijing. Human 3.6 contains motion capture data of a person captured using a high-speed 3D camera. The SEVIR and ICAR-ENSO datasets are weather and climate prediction benchmarks. The SEVIR dataset \cite{veillette2020sevir} comprises radar-derived measurements of vertically integrated liquid water (VIL) captured at 5-minute intervals with 1 km spatial resolution, serving as a benchmark for rain and hail detection. The ICAR-ENSO dataset \cite{ham2019deep} combines observational and simulation data to provide forecasts of El Niño/Southern Oscillation (ENSO), an anomaly in sea surface temperature (SST) in the Equatorial Pacific that serves as a significant predictor of seasonal climate worldwide.

\begin{table}[t!]
\begin{center}
\resizebox{0.9\columnwidth}{!}{%
\begin{tabular}{llllll}
\hline \hline
\textbf{Dataset}      & $N_{Train}$  & ${N_{Test}}$    & (C, H, W)     & $T$  & $\hat{{T}}$ \\ \hline
Moving MNIST & 10,000 & 10,000 & (1,64,64))    & 10 & 10 \\
TrafficBJ    & 19,627 & 1,334  & (2, 32, 32)   & 4  & 4  \\
Human 3.6    & 2,624  & 1,135  & (3, 128, 128) & 4  & 4  \\ \hline
SEVIR        & 35,718 & 12,159 & (1, 384, 384) & 13 & 12 \\
ICAR-ENSO    & 5,205  & 1,667  & (1,24,48)     & 12 & 14 \\ \hline \hline
\end{tabular}%
}
\end{center}
\caption{Statistics of three common benchmark datasets and two weather$\slash$climate forecasting benchmark datasets.}
\label{tab:dataset}
\end{table}

\paragraph{Evaluation metric}
For the evaluation of common benchmark datasets, we adopt widely used evaluation metrics, including MSE, MAE, Peak Signal to Noise Ratio (PSNR), and Structural Similarity Index Measure (SSIM). For rain forecasting models, we use the Critical Success Index (CSI) as an evaluation metric \cite{shi2015convolutional}. In addition, we validate ENSO forecasting using the Nino SST indices \cite{gao2022earthformer}. Specifically, the Nino3.4 index represents the averaged SST anomalies across a specific Pacific region (170$^{\circ}$W-120$^{\circ}$W, 5$^{\circ}$S-5$^{\circ}$N), and defines El Niño/La Niña events based on the SST anomalies around the equator.

\paragraph{Implementation details}
We use the Adam optimizer \cite{kingma2014adam} with $\beta_1=0.9$ and $\beta_2 = 0.999$, and a cosine scheduler without warm-up \cite{loshchilov2016sgdr} in all experiments.
The learning rate is set to 0.001 and the mini-batch size is 16.
To mitigate the learning instability of the model, we apply an exponential moving average (EMA)~\cite{rombach2022high}, which is not typically used in existing future frame prediction models.
The EMA model is updated every 10 iterations, and EMA model updates begin at 2000 iterations.
The EMA model update momentum is set to 0.995.
The Moving MNIST dataset is trained for 10K epochs, and all other datasets are trained for 2K epochs.

\subsection{Common Benchmark Results}
\paragraph{Moving MNIST}
The first row of ~\cref{tab:common} shows the performance comparison results of MAM4WF with recent leading approaches on the Moving MNIST dataset.
PhyDNet and SimVP are models that achieve state-of-the-art results in autoregressive and non-autoregressive methods, respectively.
As shown in ~\cref{tab:common}, MAM4WF achieves MSE \textbf{15.3}, MAE \textbf{51.6}, and SSIM \textbf{0.965} on the Moving MNIST dataset, outperforming both autoregressive and non-autoregressive methods on this rule-based synthetic dataset.

\paragraph{TrafficBJ}
The second row of ~\cref{tab:common} presents the performance comparison results of MAM4WF and other approaches on the TrafficBJ dataset.
In most cases of the TrafficBJ dataset, the past and future have a linear relationship, resulting in saturated values for the MSE, MAE, and SSIM of all approaches.
Nonetheless, MAM4WF achieves state-of-the-art results on the TrafficBJ dataset by a large margin.
These experimental results indicate that MAM4WF is also effective on real-world datasets with relatively easy linear relationships.

\paragraph{Human 3.6}
The third row of ~\cref{tab:common} displays the performance of MAM4WF on the Human 3.6 dataset.
Human 3.6 has a non-linear relationship between the past and future because it has to predict the future behavior of humans.
As shown in the table, MAM4WF outperforms previous models, with a 19.0 performance improvement in the MSE metric.
These experimental results demonstrate that MAM4WF is particularly robust on real-world datasets that require non-linear modeling.
\subsection{Weather$\slash$Climate Benchmark Results}
\paragraph{SEVIR}
SEVIR data has a higher resolution than MovingMNIST, and the shape of water objects is less distinct than that of numbers and humans.
Although Earthformer had already achieved state-of-the-art on this dataset, our model achieved better scores, with a CSI-M of \textbf{0.4607} and MSE of \textbf{2.9371} compared to Earthformer, as shown in ~\tabref{tab:weather}. The CSI index is generally used to validate the accuracy of the model precipitation forecast and is defined as CSI = Hits / (Hits+Misses+F.Alarms). The CSI-160 and CSI-16 calculate Hits(obs = 1, pred = 1), Misses(obs=1, pred=0), and False Alarms(obs = 0, pred = 1) based on binary thresholds from 0-255 pixel values. We used thresholds [16, 74, 133, 160, 181, 219], and the CSI-M is their averaged value \cite{gao2022earthformer}.

\paragraph{ICAR-ENSO}
The forecast evaluation, as C-Nino3.4, is calculated using the correlation skill of the three-month-averaged Nino3.4 index. We forecast up to 14-month SST anomalies (2 months more than input data for calculating three-month-averaging) from the 12-month SST anomaly observation. As shown in 

Table~\ref{tab:weather}, IAM4VP outperforms all other methods on all metrics. The C-Nino3.4-M is the mean of C-Nino3.4 over 12 forecasting steps, and C-Nino3.4-WM is the time-weighted mean correlation skill. We also computed MSE to evaluate the spatio-temporal accuracy between prediction and observation. Our model also improved the C-Nino3.4 indexes, but what is noteworthy is that the MSE is reduced by over 1 compared to other models.

\begin{table}[t!]
\begin{center}
\resizebox{\columnwidth}{!}{%
\begin{tabular}{l|c|c|c}
\hline \hline
\multicolumn{1}{c|}{\textbf{Component}} & \textbf{\#Param .(M)} & \textbf{MSE} & \textbf{Training Epoch} \\ \hline
SimVP                          &    20.4     & 23.5    & 2K             \\
+Improved Autoencoder          &    20.4     &  23.0   & 2K             \\
+ConvNeXt (ST)                 &    20.4     &  18.2   & 2K             \\ \hline
+STR                           &    20.5     &   17.6  & 2K             \\ 
+STR                           &    20.5     &   \textbf{16.2}  & 10K             \\ \hline
+Time Step MLP Embedding       &   20.6      &   19.3  & 2K            \\ 
+Time Step MLP Embedding       &   20.6      &   \textbf{17.4}  & 10K            \\ \hline
+Masked Autoregressive  w/o $M_{g}$    &    20.7     &   25.2  & 10K          \\
+Masked Autoregressive  w/  $M_{g}$   &    20.7     &   16.9  & 2K            \\
+Masked Autoregressive  w/  $M_{g}$   &    20.7     &   \textbf{15.8}  & 10K           \\ \hline
+Learned Prior (MAM4WF)                       &  20.7       &   15.9  & 2K+Fine-tune \\
+Learned Prior (MAM4WF)                       &  20.7       &   \textbf{15.3}  & 10K+Fine-tune 
\\ \hline \hline
\end{tabular}%
}
\end{center}
\caption{Performance analysis for MAM4WF components.}
\label{tab:able1}
\end{table}

\subsection{Ablation Study}
\paragraph{Component effect}
We conducted an ablation study on the Moving MNIST dataset to analyze the effect of each component of MAM4WF. The results of this study are presented in \tabref{tab:able1}, where we analyzed the performance of MAM4WF by sequentially adding each component to the SimVP baseline.

In SimVP, we observed a performance improvement of 0.5 by changing the encoder structure from (Conv, GroupNorm, LeakyReLU) to (Conv, LayerNorm, SiRU) and the decoder structure from (ConvTranspose, GroupNorm, LeakyReLU) to (Conv, LayerNorm, SiRU, PixelShuffle).
Furthermore, by changing the spatio-temporal predictor from the existing inception block to the ConvNeXt block, we observed a significant performance improvement of +4.8.
These experimental results indicate the importance of the spatio-temporal predictor in the video frame prediction.

Next, we applied STR~\cite{seo2022simple}, a refinement module that demonstrated good performance on weather forecast datasets, resulting in a performance improvement of 0.6. By changing the model from the MIMO to the implicit MISO and extending the training schedule from 2K to 10K epochs, we achieved an additional performance boost of +0.2. While this improvement may seem small, it was crucial to implementing the stacked autoregressive model structure.

\begin{figure}[t!]
    \centering
    \includegraphics[width=1.0\columnwidth]{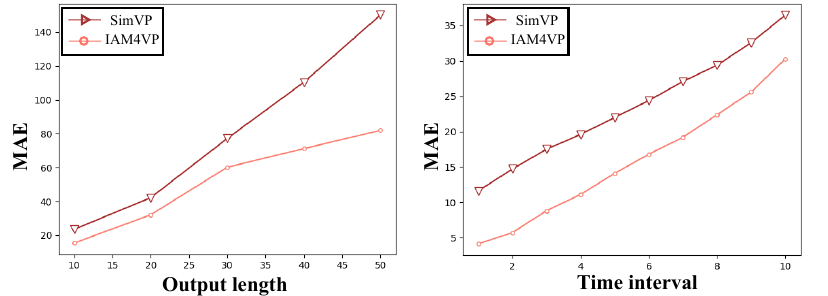}
    \caption{Performance comparison experiment according to output length and time interval changes of SimVP and MAM4VP on the Moving MNIST dataset.}
    \label{fig:output}
\end{figure}

\paragraph{Output condition dependency}
The experimental results of our analysis on the performance according to the output condition are presented in ~\cref{fig:output}.
The left graph shows a performance comparison between SimVP and MAM4WF according to the output length.
As shown in the figure, SimVP exhibits a linearly increasing error as the output length increases, whereas MAM4WF does not, indicating that MAM4WF is more robust to the output length.

The right graph shows the performance analysis result according to time steps, where the error in both methods increases as the time step increases, as expected.
These experimental results demonstrate that predicting long output lengths is more challenging, which increases the difficulty of the prediction.
Overall, the experimental results suggest that MAM4WF  outperforms SimVP when predicting long output lengths.
Specifically, error of MAM4WF increases linearly up to output length 20 (i.e., time step 20), but does not increase linearly beyond this point.
These findings underscore the importance of the MAM4WF design, and the components of this model can be usefully applied in tasks such as weather forecasting.

\section{Qualitative Results}
\begin{figure*}[t!]
    \centering
    \includegraphics[width=2.0\columnwidth]{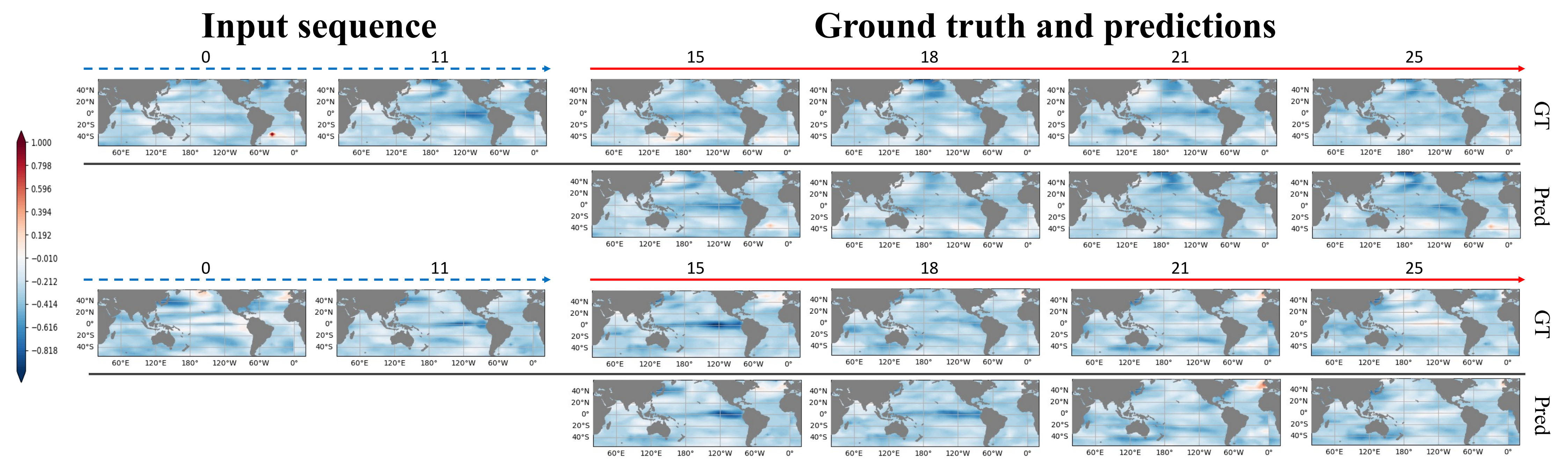}
    \caption{Prediction results of MAM4WF on the ICRA-ENSO dataset. The color bar means SST anomalies on the global map. Best viewed with zoom.}
    \label{fig:ffi}
\end{figure*}
\begin{figure}[t!]
     \centering
     \includegraphics[width=0.45\textwidth]{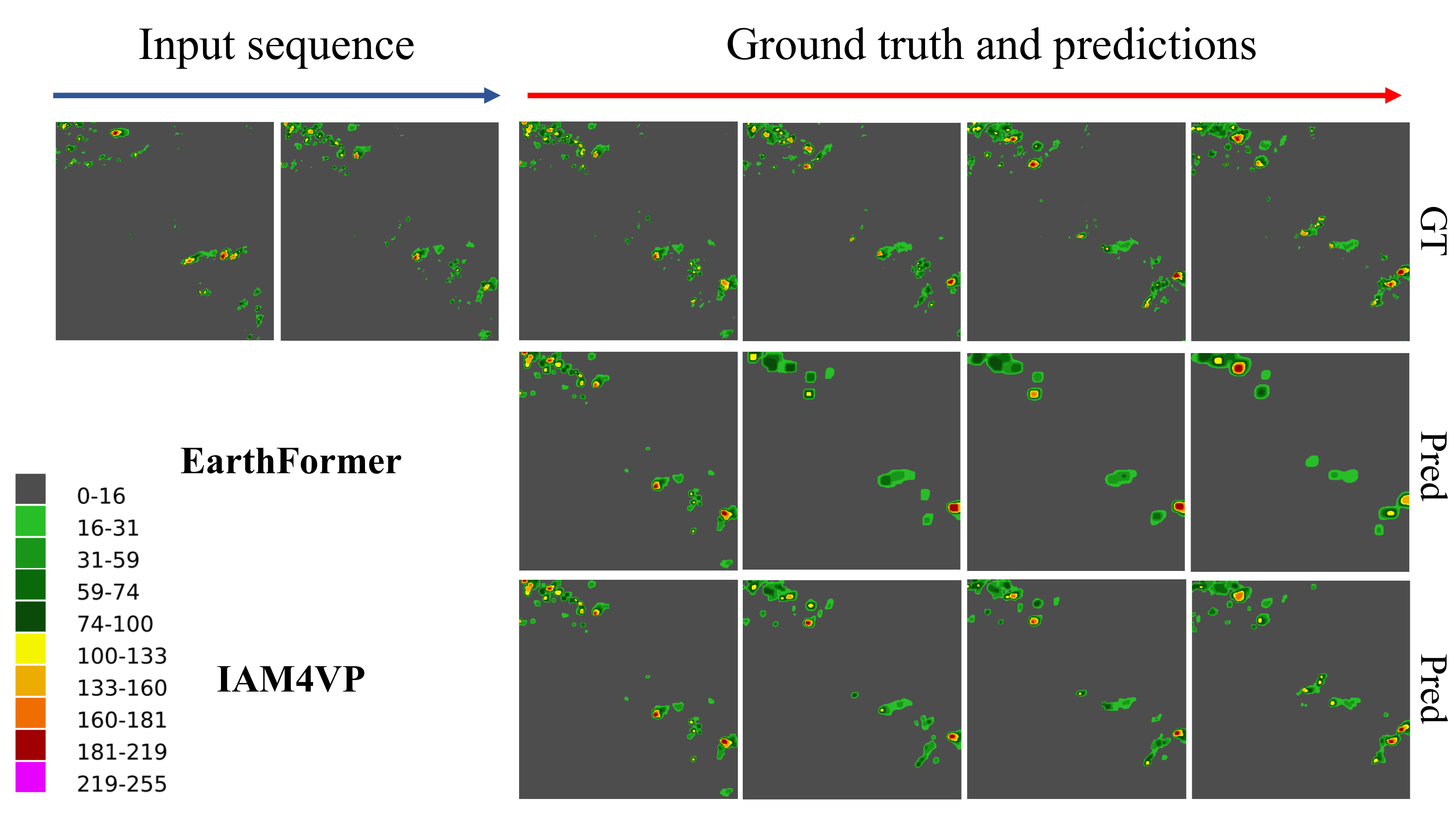}
     \caption{Prediction results of MAM4WF and Earthformer on the SEVIR dataset, represented by vertically integrated liquid water contents (0-255 scale) shown on the color bar.}
     \label{fig:seivr}
\end{figure}

\paragraph{ICRA-ENSO} The predicted results for the ICAR-ENSO dataset are depicted in ~\cref{fig:ffi}, showcasing the global map of SST anomalies.
In the color bar, blue denotes negative SST anomalies, while red represents positive ones.
The results indicate that MAM4WF tends to overestimate values but successfully predicts anomalous patterns transitioning from negative to positive, in line with observations (GT). 
Notably, even if the model produces incorrect predictions in the initial stages for certain regions, subsequent predictions align with the correct SST anomaly pattern. This means that the MAM4WF model does not carry forward early errors to subsequent predictions.
\paragraph{SEVIR}
~\cref{fig:seivr} displays the predicted outcomes of MAM4WF and Earthformer on the SEVIR dataset. This dataset signifies radar-estimated liquid water in a vertical air column, with the color bar representing water mass per unit area.
A higher VIL value is indicative of elevated water content, increasing the likelihood of precipitation or even the occurrence of hail during storms.
From 12 input images (the two on the left), we forecast 12 future frames (four samples on the right). In both examples, the initial future frame predictions by both models seem plausible. However, disparities begin to emerge from frame 17 onward.
Earthformer exhibits issues with blurriness, characterized by oversimplified cell boundaries and the omission of smaller cells. This is a widespread limitation of deep-learning-based weather prediction.
Conversely, MAM4WF offers precise predictions of cell boundaries up to frame 24.
Notably, even the smaller cells are retained throughout training, suggesting that some cells, which are not apparent in Earthformer's outputs, are discernible in our model's predictions.
\section{Limitation and Future Work}
Unlike the existing autoregressive model, the masked autoregressive method lacks flexibility for lead time.
For example, the lead time of the existing autoregressive model can be increased as much as desired by adjusting the autoregressive step even if the performance is degraded.
However, since the masked autoregressive method has to fix the length of the feature map input to the model, the length of the lead time cannot exceed the fixed length.
However, the masked autoregressive method can also perform inference in the same way as the existing autoregressive methods, but this does not match the motivation of the masked autoregressive method.
In our future work, we will conduct research on increasing the output length flexibility of the masked autoregressive method.

\section{Conclusion}
In this paper, we introduce a novel masked autoregressive model for weather forecasting, MAM4WF, which combines the strengths of both autoregressive and lead time embedding methods using a masked autoregressive approach.
By integrating the lead time embedding method and maintaining an error-free queue, MAM4WF effectively addresses the error accumulation problem that has hindered traditional autoregressive methods.
Moreover, the incorporation of a masked autoregressive architecture and an error-aware queue ensures that MAM4WF consistently captures the essential spatiotemporal correlations between predictions — a crucial aspect of atmospheric phenomena often overlooked by previous lead time embedding models.
Extensive experiments demonstrate that MAM4WF achieves state-of-the-art performance across two weather and climate forecasting datasets, as well as three video predictions.

{\small
\bibliographystyle{ieee_fullname}
\bibliography{egbib}
}

\end{document}